\documentclass[10pt,twocolumn,letterpaper]{article}
\def\cvprPaperID{2759}
\def\confName{CVPR}
\def\confYear{2023}

\def\OADP{Object-Aware Distillation Pyramid}
\def\OAKE{Object-Aware Knowledge Extraction}
\def\DP{Distillation Pyramid}

\makeatletter
\newcommand{\@mAP}[2]{mAP$^{\text{#1}}_{#2}$}
\newcommand{\@AP}[1]{AP\textsubscript{#1}}
\newcommand{\mAPN}[1]{\@mAP{N}{#1}}
\newcommand{\mAPB}[1]{\@mAP{B}{#1}}
\newcommand{\mAPPL}[1]{\@mAP{PL}{#1}}
\newcommand{\mAP}[1]{\@mAP{}{#1}}
\def\APr{\@AP{r}}
\def\APc{\@AP{c}}
\def\APf{\@AP{f}}
\def\AP{\@AP{}}
\makeatother

\newif\ifarxiv
\arxivtrue

\ifarxiv
  \usepackage[pagenumbers]{cvpr}
\else
  \usepackage{cvpr}
\fi

\usepackage{bbm}
\usepackage{booktabs}
\usepackage{graphicx}
\usepackage{subcaption}
\usepackage[table,xcdraw]{xcolor}
\usepackage[pagebackref,breaklinks,colorlinks]{hyperref}

\ifarxiv
  \makeatletter
  \@namedef{ver@everyshi.sty}{}
  \makeatother
\fi
\usepackage{utils}

\AtBeginEnvironment{equation}{\small}
\AtBeginEnvironment{eqnarray}{\small}

\begin{document}

\renewcommand{\paragraph}[1]{\vspace{6pt minus 6pt}\noindent\textbf{#1}}

\title{\OADP{} for Open-Vocabulary Object Detection}

\ifarxiv
  \author{
    Luting Wang$^1$\quad
    Yi Liu$^1$\quad
    Penghui Du$^1$\quad
    Zihan Ding$^1$\quad
    Yue Liao$^1$\thanks{Corresponding author (liaoyue.ai@gmail.com)}\\
    Qiaosong Qi$^2$\quad
    Biaolong Chen$^2$\quad
    Si Liu$^1$\\
    $^1$Institute of Artificial Intelligence, Beihang University\quad
    $^2$Alibaba Group
  }
\else
  \author{
    Luting Wang$^{1,3}$\quad
    Yi Liu$^{1,3}$\quad
    Penghui Du$^{1,3}$\quad
    Zihan Ding$^{1,3}$\quad
    Yue Liao$^{1,3}$\thanks{Corresponding author (liaoyue.ai@gmail.com)}\\
    Qiaosong Qi$^2$\quad
    Biaolong Chen$^2$\quad
    Si Liu$^{1,3}$\\
    $^1$Institute of Artificial Intelligence, Beihang University\quad
    $^2$Alibaba Group\\
    $^3$Hangzhou Innovation Institute, Beihang University
  }
\fi

\maketitle

\begin{abstract}
  Open-vocabulary object detection aims to provide object detectors trained on a fixed set of object categories with the generalizability to detect objects described by arbitrary text queries.
  Previous methods adopt knowledge distillation to extract knowledge from Pretrained Vision-and-Language Models (PVLMs) and transfer it to detectors.
  However, due to the non-adaptive proposal cropping and single-level feature mimicking processes, they suffer from information destruction during knowledge extraction and inefficient knowledge transfer.
  To remedy these limitations, we propose an \OADP{} (OADP) framework, including an \OAKE{} (OAKE) module and a \DP{} (DP) mechanism.
  When extracting object knowledge from PVLMs, the former adaptively transforms object proposals and adopts object-aware mask attention to obtain precise and complete knowledge of objects.
  The latter introduces global and block distillation for more comprehensive knowledge transfer to compensate for the missing relation information in object distillation.
  Extensive experiments show that our method achieves significant improvement compared to current methods.
  Especially on the MS-COCO dataset, our OADP framework reaches $35.6$ \mAPN{50}, surpassing the current state-of-the-art method by $3.3$ \mAPN{50}.
  Code is released at \url{https://github.com/LutingWang/OADP}.
\end{abstract}

\section{Introduction}

\begin{figure}[t]
  \centering
  \begin{subfigure}{0.8\linewidth}
    \includegraphics[width=\textwidth]{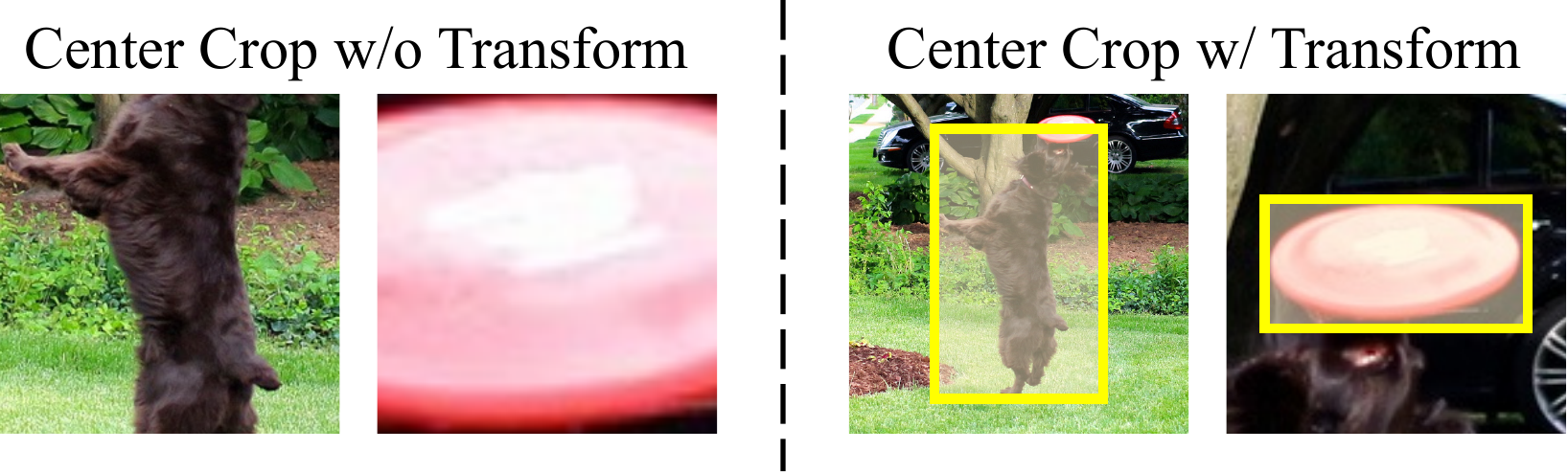}
    \caption{Knowledge Extraction}
  \end{subfigure}
  \begin{subfigure}{0.8\linewidth}
    \includegraphics[width=\linewidth]{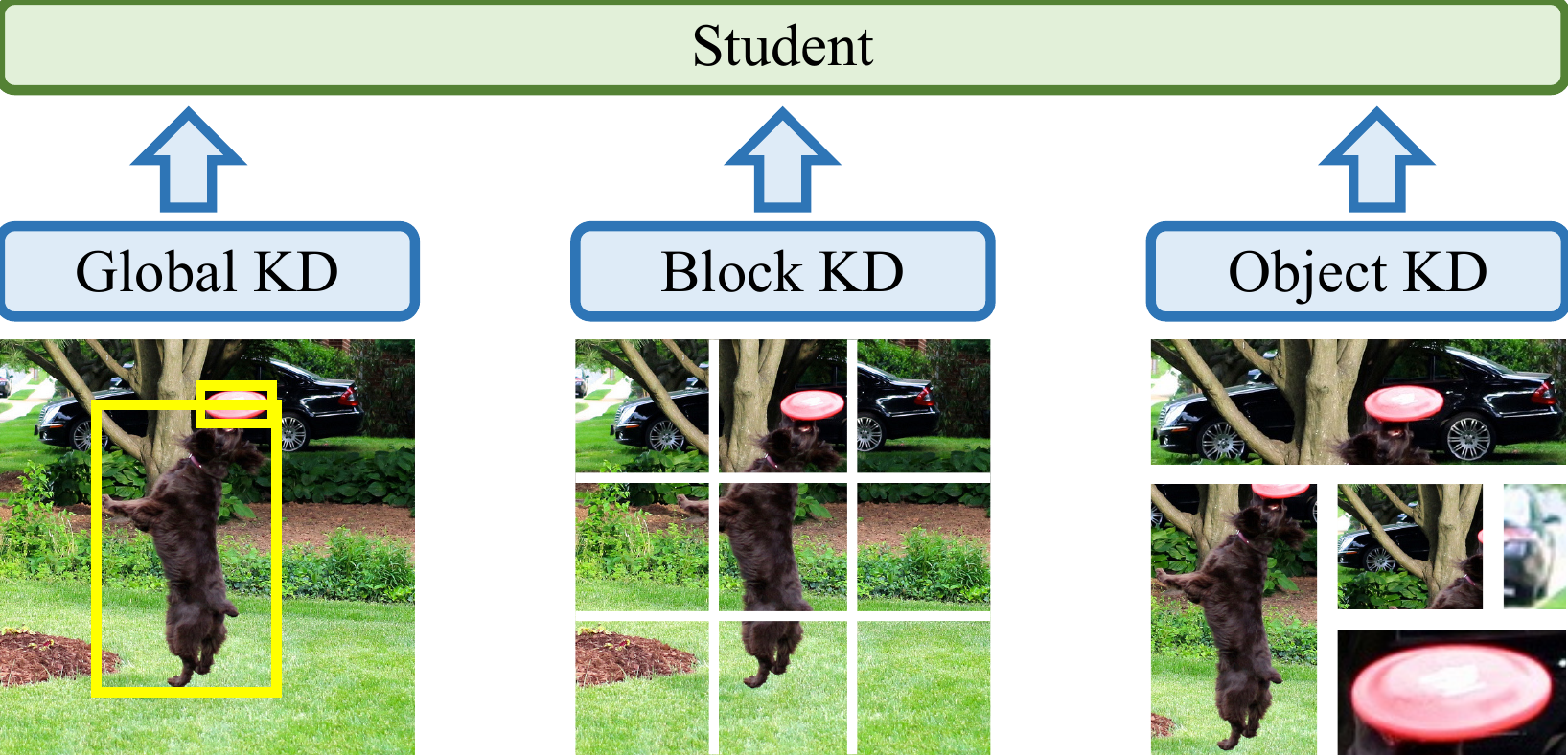}
    \caption{Knowledge Transfer}
  \end{subfigure}
  \caption{
    An overview of our OADP framework.
    (a) Directly applying center crop on proposals may throw informative object parts away, resulting in ambiguous image regions.
    In contrast, our OAKE module extracts complete objects and reduces the influence of surrounding distractors.
    (b) Our DP mechanism includes global, block, and object KD to achieve effective knowledge transfer.
  }
  \label{fig:fig1}
\end{figure}

Open-vocabulary object detection (OVD)~\cite{ovr_cnn} aims to endow object detectors with the generalizability to detect \textit{open categories} including both \textit{base} and \textit{novel} categories where only the former are annotated in the training phase.
Pretrained Vision-and-Language Models~(PVLMs, \eg, CLIP~\cite{clip} and ALIGN~\cite{align}) have witnessed great progress in recent years, and Knowledge Distillation~(KD)~\cite{kd} has led to a wave of unprecedented advances transferring the zero-shot visual recognition ability from PVLMs to detectors~\cite{vild, hierkd, owl_vit, ov_detr, regionclip, gen_vlkt}.
KD typically comprises two essential steps, \ie, \textit{knowledge extraction} and then \textit{knowledge transfer}.
A common practice in OVD is to crop objects with class-agnostic proposals and use the teacher (\eg, CLIP visual encoder) to extract knowledge of the proposals.
The knowledge is then transferred to the detector (\eg, Mask R-CNN~\cite{mask_rcnn}) via feature mimicking.

Despite significant development, we argue that conventional approaches still have two main limitations:
1) \textit{Dilemma between comprehensiveness and purity during knowledge extraction}.
As proposals have diverse aspect ratios, the fixed center crop strategy to square them may cut out object parts (\cref{fig:fig1} (a)).
Enlarging those proposals via resizing function may alleviate this problem, but additional surrounding distractors may confuse the teacher to extract accurate proposal knowledge.
2) \textit{Missing global scene understanding during knowledge transfer}.
Conventional approaches merely concentrate on object-level knowledge transfer by directly mimicking the teacher's features of individual proposals.
As a result, the student cannot fully grasp the contextual characteristics  describing the interweaving of different objects.
In light of the above discussions, we propose an \OADP{} (OADP) framework to excavate the teacher's knowledge accurately and effectively transfer the knowledge to the student.

To preserve the complete information of proposals while extracting their CLIP image embeddings,
we propose an \OAKE{} (OAKE) module.
Concretely, given a proposal, we square it with an adaptive resizing function to avoid destroying the object structure and involve object information as much as possible.
However, the resizing process inevitably introduces environmental context, which may contain some distractors that confuse the teacher.
Therefore, we propose to utilize an object token \texttt{[OBJ]} whose interaction manner during the forward process is almost the same as the class token \texttt{[CLS]} except that it only attends to patch tokens covered by the original proposal.
In this way, the extracted embeddings contain precise and complete
knowledge of the proposal object.


To facilitate complete and effective knowledge transfer, we propose a \DP (DP) mechanism (\cref{fig:fig1} (b)).
As previous works only adopt object distillation to align the feature space of detectors and PVLMs, the relation between different objects is neglected.
Therefore, we propose global and block distillation to compensate for the missing relation information in object distillation.
For global distillation, we optimize the $\mathcal{L}_1$ distance between the detector backbone and the CLIP visual encoder so that the detector learns to encode rich semantics implied in the image scene.
However, the CLIP visual encoder is prone to ignore background information, which may also be valuable for detection.
Therefore, we take a finer step to divide the input image into several blocks and optimize the $\mathcal{L}_1$ distance between the block embeddings of the detector and the CLIP image encoder.
Overall, the above three distillation modules constitute a hierarchical distillation pyramid, allowing for the transfer of more diversified knowledge from CLIP to the detectors.

We demonstrate the superiority of our OADP framework on MS-COCO~\cite{coco} and LVIS~\cite{lvis} datasets.
On MS-COCO, it improves the state-of-the-art results of \mAPN{50} from $32.3$ to $35.6$.
On the LVIS dataset, our OADP framework reaches $21.9$ \APr{} on the object detection task and $21.7$ \APr{} on the instance segmentation task, leading the former methods by more than $1.1$ \APr{} and $1.9$ \APr{} respectively.



\section{Related Work}

\paragraph{Knowledge Distillation for Object Detection.}
KD~\cite{kd, fitnets} is a technology that helps train compact student models under the supervision of powerful teacher models.
Chen \etal~\cite{detection_kd} apply KD to object detection by implementing feature-based and response-based loss for Faster R-CNN.
Li \etal~\cite{mimic} apply $\mathcal{L}_2$ loss on features sampled by student proposals.
FGFI~\cite{fgfi} only distills foreground regions near the object anchors.
DeFeat~\cite{defeat} distills the foreground and background regions simultaneously with different factors.
GID~\cite{gid} distills regions where the student and teacher perform differently.
G-DetKD~\cite{gdetkd} proposes a general distillation framework for object detectors.
FKD~\cite{fkd} distills the attention map to emphasize the changeable areas.
FGD~\cite{fgd} proposes focal and global distillation for comprehensive knowledge transfer.
Compared to these detection KD methods~\cite{mgd, takd, frs, lgd, icd}, our work concentrates on knowledge transfer from PVLMs to detectors to enable open-vocabulary detection.

\paragraph{Open-Vocabulary Detection.}
OVD~\cite{owl_vit, mdetr, x_detr, detclip} aims to train a model that can detect objects of arbitrary categories, even if the categories are not seen during training.
OVR-CNN~\cite{ovr_cnn} is the seminal work that proposes this problem and achieves great performance using image captions as well as bounding box annotations.
With the prevalence of PVLMs~\cite{clip, align}, ViLD~\cite{vild} proposes to distill the open-vocabulary knowledge from CLIP to the detector.
DetPro~\cite{detpro} improves upon ViLD with prompt optimization.
RegionCLIP~\cite{regionclip} develops a pretraining strategy to learn region-text alignment.
Detic~\cite{detic} adopts weak supervisions to jointly train the detector.
GLIP~\cite{glip} pretrains on massive image-text pairs in a self-training fashion by unifying the detection and grounding tasks.
HierKD~\cite{hierkd} proposes instance- and global-level distillation for one-stage detectors.
OV-DETR~\cite{ov_detr} turns DETR into an open-vocabulary detector with conditional binary matching.
VL-PLM~\cite{vl_plm} leverages pseudo labels on novel categories to augment the detector.
PB-OVD~\cite{pb_ovd} generates pseudo labels based on the image captions.
PromptDet~\cite{promptdet} establishes a scalable pipeline with regional prompt learning and self-training.
In this paper, we propose an OADP framework focusing on comprehensive object knowledge extraction and effective knowledge transfer.

\section{OVD Benchmarks}
\label{sec:ovd_benchmarks}

According to the training data, we summarize the existing OVD methods into four types of benchmarks: Vanilla OVD (V-OVD), Caption-based OVD (C-OVD), Generalized OVD (G-OVD), and Weakly Supervised OVD (WS-OVD).
All benchmarks rely on instance-level annotations and large-scale image-text pairs to learn OVD.
Some of them use more types of data, as shown in \cref{tab:benchmarks}.
For clarity, we define base categories as those included in the instance-level annotations, and novel categories are the others.

\begin{table}[t]
  \centering
  \resizebox{0.85\linewidth}{!}{%
    \begin{tabular}{r|ccc}
      \toprule
      Benchmark & Caption    & Category Prior & Image Label \\ \midrule
      V-OVD     &            &                &             \\
      C-OVD     & \checkmark &                &             \\
      G-OVD     &            & \checkmark     &             \\
      WS-OVD    & \checkmark & \checkmark     & \checkmark  \\ \bottomrule
    \end{tabular}%
  }
  \caption{
    Summary of OVD benchmarks.
    ``Caption'': in-domain captions like COCO-Captions.
    ``Category Prior'': human priors on novel categories.
    ``Image Label'': image-level category labels.
  }
  \label{tab:benchmarks}
\end{table}

\paragraph{V-OVD}~\cite{vild, detpro, regionclip, glip, mdetr, x_detr, owl_vit, detclip} is a pure OVD benchmark setting, which requires the detector only to train in an object detection dataset with fixed categories set.
Any information about the novel categories is unavailable, but unannotated data is allowed.
A common practice for this benchmark is to learn open vocabulary knowledge from image-text pairs and transfer the knowledge to detectors through transfer learning or knowledge distillation.
V-OVD is similar to ZSD~\cite{zsd, contrast_zsd, zs_yolo, transductive_zsd}, except that V-OVD relies on large-scale image-text pairs to acquire open-vocabulary knowledge.
Recently, V-OVD has attracted more and more researchers with the development of PVLMs.

\paragraph{C-OVD}~\cite{hierkd, ovr_cnn, locov, pb_ovd} adds additional image caption annotation to the V-OVD benchmark.
Note that by image caption data, we refer to the in-domain captions of the instance-level annotations, \eg, COCO-Captions~\cite{coco_captions}, instead of the large-scale image-text pairs, \eg, CC3M~\cite{cc} and CLIP400M~\cite{clip}.
The in-domain captions enrich the instance-level annotations and imply a distribution of potential novel categories.
Compared with the V-OVD benchmark, C-OVD requires slightly more annotations and is expected to perform better.

\paragraph{G-OVD}~\cite{ov_detr, vl_plm, promptdet} introduces human priors on novel categories to the V-OVD benchmark.
Intuitively, if some novel categories are far more likely to appear during inference, it would be beneficial to prepare for them during training.
Most existing methods assume that all the dataset's category names (including the novels) are known to the detectors during training.
Therefore, the performance of G-OVD methods may not be fairly comparable with V-OVD and C-OVD methods.
A typical solution is to generate instance-level pseudo annotations for the categories.

\paragraph{WS-OVD}~\cite{detic} further takes advantage of image-level category labels beyond G-OVD.
Similar to Weakly Supervised Detection (WSD)~\cite{wsdnn, cap2det}, the image-level category labels reflect the presence of the base and novel categories in each image.
Thus, the annotation cost is far more than the benchmarks above.
In this case, WS-OVD methods have the greatest potential to push the limit of OVD further.

\section{\OADP{}}

We first briefly review the task definition of OVD and the architecture of Faster R-CNN in \cref{sec:preliminaries}.
Then, we present the overview of our OADP framework in \cref{sec:overview_of_oadp}.
\Cref{sec:object_distillation} and \cref{sec:global_and_block_distillation} introduce the OAKE module and the DP mechanism in detail.
Finally, in \cref{sec:pseudo_label_generation}, we demonstrate the procedure to generate pseudo labels based on OAKE.

\subsection{Preliminaries}
\label{sec:preliminaries}

We represent traditional object detection datasets as $\mathcal{D} = \{(\mathbf{I}_i, \mathcal{O}_i)\}_{i = 1}^{|\mathcal{D}|}$, where $\mathbf{I}_i$ is the $i$-th image and $\mathcal{O}_i = \{o_{ij}\}_{j = 1}^{|\mathcal{O}_i|}$ is the corresponding set of annotated objects.
Each object $o$ is a pair of object bounding box $b \in \mathbb{R}^4$ and category $y \in \mathcal{C}$, where $\mathcal{C}$ is the category space of the dataset.
We denote the training and validation datasets as $\mathcal{D}^T$ and $\mathcal{D}^V$, respectively.

By the convention of OVD, we refer to the category space of $\mathcal{D}^T$ and $\mathcal{D}^V$ as $\mathcal{C}^B$ and $\mathcal{C}$ respectively.
Normally, $\mathcal{C}^B \subset \mathcal{C}$.
Categories in $\mathcal{C}^B$ are called base categories, and those that only appear in $\mathcal{D}^V$ are called novel categories.
The novel category space is denoted as $\mathcal{C}^N = \mathcal{C} \setminus \mathcal{C}^B \ne \varnothing$.
For each category $c \in \mathcal{C}$, we use a pretrained text encoder $\mathcal{T}$ to encode its semantic embedding $t_c \in \mathbb{R}^d$.
Specifically, we use a trainable embedding $t_{\text{bg}} \in \mathbb{R}^d$ to represent the background class \texttt{bg}.

Since our work is based on Faster R-CNN~\cite{faster_rcnn}, we briefly recap its framework.
Given an image $\mathbf{I} \in \mathbb{R}^{H \times W \times 3}$, the backbone (including FPN~\cite{fpn}) encodes a set of hierarchical feature maps $\mathcal{F} = \{\mathbf{F}_2, \mathbf{F}_3, \cdots, \mathbf{F}_6\}$ and the Region Proposal Network (RPN) generates a set of proposals $\mathcal{P} \subset \mathbb{R}^{4}$.
Then the R-CNN head performs RoI Align on $\mathcal{F}$ to extract proposal embeddings $\mathcal{E} = \{e_p\}_{p \in \mathcal{P}} \subset \mathbb{R}^{d}$.
The logit of a proposal $p$ being of category $c$ can be defined as:
\begin{equation}
  l(p, c) = \frac{e_p \cdot t_c}{\Vert e_p\Vert \cdot \Vert t_c\Vert},
\end{equation}
where $\cdot$ is the dot product, $t_c$ is the category embedding of $c$.
For simplicity, we ignore the temperature $\tau$ in CLIP~\cite{clip}.
The probability of a proposal $p$ belonging to category $c \in \mathcal{C} \cup \{\texttt{bg}\}$ is:
\begin{equation}
  P_{\mathcal{C}}(p, c) = \frac
  {\exp\left(l(p, c)\right)}
  {\sum\limits_{c' \in C \cup \{\texttt{bg}\}} \exp\left(l(p, c')\right)}.
\end{equation}

During training, each proposal $p$ is assigned a category label $y_p \in \mathcal{C}^B \cup \{\texttt{bg}\}$.
The R-CNN loss is defined as:
\begin{equation}
  \label{eq:rcnn_loss}
  \mathcal{L} = -\sum_{p \in P} \log P_{\mathcal{C}^B}(p, y_p).
\end{equation}
For simplicity, we ignore the regression term in $\mathcal{L}$.

\subsection{Overview of OADP}
\label{sec:overview_of_oadp}

\begin{figure*}[t]
  \centering
  \includegraphics[width=0.9\linewidth]{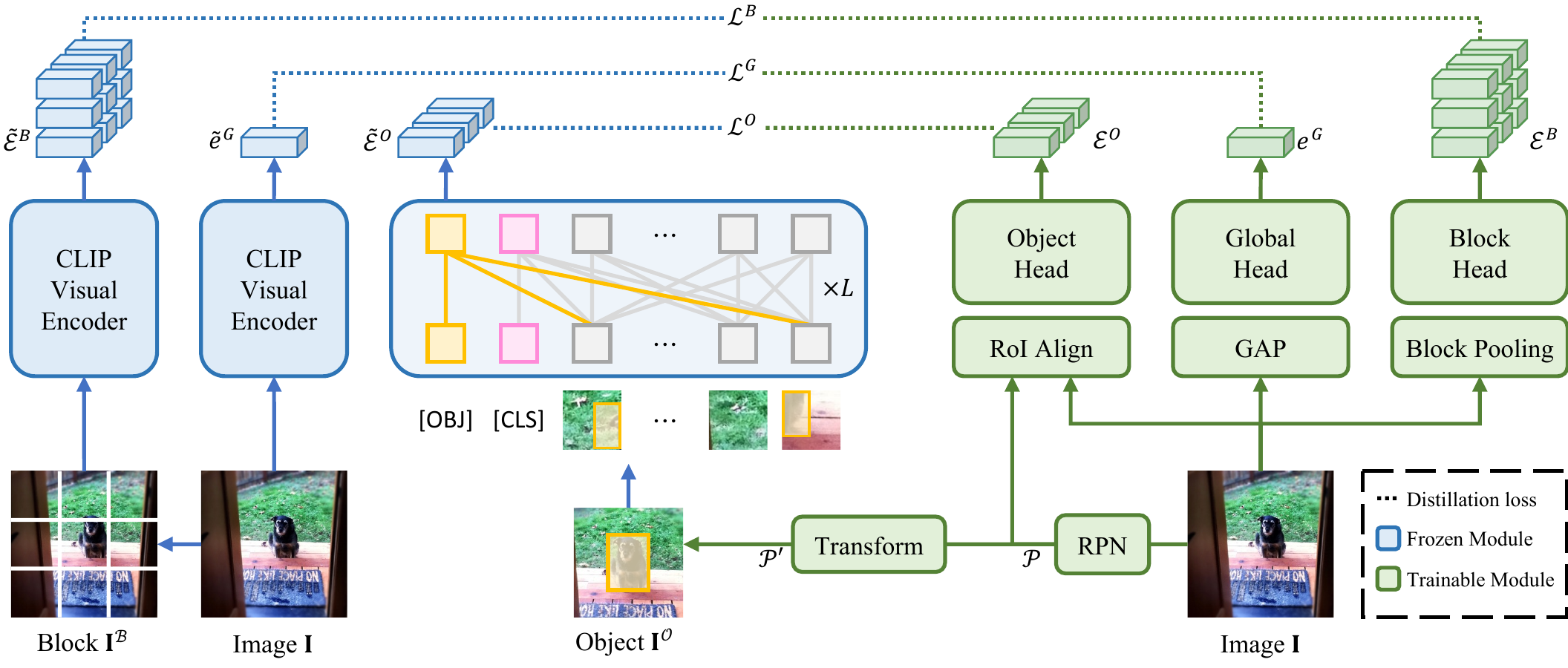}
  \caption{
  Illustration of our OADP training pipeline.
  We adopt a pyramid architecture comprising three distillation modules: global, block, and object.
  Given an image $\mathbf{I}$, RPN generates proposals $\mathcal{P}$.
  For object distillation, RoI Align and Object Head are applied for proposal embeddings $\mathcal{E}^O$.
  To extract complete and pure object knowledge from CLIP, we crop the image regions $\mathbf{I}^{\mathcal{O}}$ based on the transformed proposals $\mathcal{P}'$ and feed them to $L$ layers of masked attention, where an extra \texttt{[OBJ]} token (yellow) attends to the patches covered by the original proposal.
  For global and block distillation, GAP and block pooling are used before the corresponding heads to extract the global and block embeddings ($\mathcal{E}^B$ and $e^G$).
  The teacher embeddings $\tilde{\mathcal{E}}^B$ and $\tilde{e}^G$ are extracted via CLIP from $\mathbf{I}$ and $\mathcal{I}^B$ respectively.
  }
  \label{fig:framework}
\end{figure*}

To inject open-vocabulary concepts into Faster R-CNN, we propose an \OADP{} (OADP) framework (\cref{fig:framework}), which first extracts knowledge from CLIP~\cite{clip} and then transfers it to the detector through knowledge distillation (KD)~\cite{kd}.
Specifically, we propose an \OAKE{} (OAKE) module, which inserts an \texttt{[OBJ]} token into the frozen CLIP visual encoder $\mathcal{V}$ to extract informative knowledge from expanded region proposals selectively.
For more effective knowledge transfer, we propose a \DP (DP) mechanism comprising an object distillation module $\mathcal{M}^O$, a block distillation module $\mathcal{M}^B$, and a global distillation module $\mathcal{M}^G$.
The losses of the three modules are denoted as $\mathcal{L}^O$, $\mathcal{L}^B$, and $\mathcal{L}^G$, respectively.
The total training loss is:
\begin{equation}
  \mathcal{L}^{\text{all}} = \mathcal{L} + w^O \cdot \mathcal{L}^O + w^B \cdot \mathcal{L}^B + w^G \cdot \mathcal{L}^G,
\end{equation}
where $\mathcal{L}$ is the R-CNN loss as defined in \cref{eq:rcnn_loss}; $w^O$, $w^B$, and $w^G$ are loss weights.

We follow the inference pipeline of ViLD-ensemble~\cite{vild} and use $\mathcal{M}^O$ to calibrate $P_{\mathcal{C}}(p, c)$.
Similar to the R-CNN head, $\mathcal{M}^O$ extracts the proposal embeddings $\mathcal{E}^O = \{e^O_p\}_{p \in \mathcal{P}} \subset \mathbb{R}^{d}$ and computes the logits:
\begin{eqnarray}
  l^O(p, c) &=& \frac{e^O_p \cdot t_c}{||e^O_p|| \cdot ||t_c||},\\
  P^O_{\mathcal{C}}(p, c) &=& \frac
  {\exp\left(l^O(p, c)\right)}
  {\sum\limits_{c' \in C} \exp\left(l^O(p, c')\right)},
\end{eqnarray}
where $t_c$ is the category embedding of $c$.
The calibrated probability $P_{\mathcal{C}}^{\text{cal}}(p, c)$ is:
\begin{equation}
  P_{\mathcal{C}}^{\text{cal}}(p, c) = \begin{cases}
    \left(P_{\mathcal{C}}(p, c)\right)^{\lambda} \cdot \left(P^O_{\mathcal{C}}(p, c)\right)^{(1 - \lambda)}, & c \in \mathcal{C}^B \\
    \left(P_{\mathcal{C}}(p, c)\right)^{(1 - \lambda)} \cdot \left(P^O_{\mathcal{C}}(p, c)\right)^{\lambda}, & c \in \mathcal{C}^N \\
    1 - \sum_{c' \in \mathcal{C}} P_{\mathcal{C}}(p, c'),                                                    & c = \texttt{bg}
  \end{cases}
\end{equation}
where $\lambda$ is set to $2 / 3$.
Note that the block and global distillation modules are not used during the inference phase, so the computation cost of our OADP framework is the same as ViLD-ensemble.

\subsection{Object Distillation}
\label{sec:object_distillation}

The object distillation module $\mathcal{M}^O$ aims to transfer the object-level knowledge from CLIP~\cite{clip} to the detector.
For each proposal $p \in \mathcal{P}$, $\mathcal{M}^O$ motivates the detector to extract a proposal embedding $e^O_p$ that resembles the corresponding embedding $\tilde e^O_p$ extracted by the CLIP visual encoder $\mathcal{V}$:
\begin{equation}
  \mathcal{L}^O = \mathcal{L}_1(\mathcal{E}^O, \tilde{\mathcal{E}}^O),
\end{equation}
where $\tilde{\mathcal{E}}^O = \{\tilde e^O_p\}_{p \in \mathcal{P}} \subset \mathbb{R}^{d}$ denotes the proposal embeddings extracted by $\mathcal{V}$.
Naturally, the quality of $\tilde{\mathcal{E}}^O$ affects the accuracy of $P^O_{\mathcal{C}}(p, c)$ to a large extent.
However, current approaches only yield sub-optimal $\tilde{\mathcal{E}}^O$ due to the dilemma between information comprehensiveness and less noise.
For example, when non-square proposal regions are directly passed to CLIP, the center crop operation in $\mathcal{V}$ will cut out informative parts of an object, leading to incomplete structural knowledge about the object.
On the other hand, if the proposals are squared or enlarged, the proposal regions will contain more ambient contexts, which may corrupt the proposal embeddings.

To acquire more accurate $\tilde{\mathcal{E}}^O$, we propose an \OAKE{} (OAKE) module, where the proposals are first transformed and then encoded with a modified version of $\mathcal{V}$.
Given a proposal $p \in \mathcal{P}$, the transformed proposal $p'$ is a square with side length $s = \sqrt{r\times p_h\times p_w}$, where $r$ is a constant scale ratio, $p_h$ and $p_w$ are the height and width of $p$.
The center of $p'$ is initially the same as $p$ but may be translated if $p'$ exceeds the image boundaries.
All transformed proposals constitute $\mathcal{P}'$.

While $\mathcal{P}'$ contributes to the comprehensiveness of object knowledge, it also includes surrounding distractors.
To further suppress the contextual noise, we introduce an \texttt{[OBJ]} token into $\mathcal{V}$ and substitute original attention layers with masked attention layers.
The modified version of $\mathcal{V}$ is denoted as $\mathcal{V}'$.
The input of $\mathcal{V'}$ is a set of image regions $\mathbf{I}^{\mathcal{O}} = \{\mathbf{I}_{p'}\}_{p' \in \mathcal{P}'}$, where $p'$ is a transformed proposal and $\mathbf{I}_{p'}$ is the cropped image region of $p'$.
Same as $\mathcal{V}$, each $\mathbf{I}_{p'}$ is first mapped to a sequence of tokens $\mathbf{X} \in \mathbb{R}^{N_x \times d_x}$, where $\mathbf{X}_{1:N_x - 1}$ are the patch tokens of $\mathbf{I}_{p'}$ and $\mathbf{X}_{N_x}$ is the \texttt{[CLS]} token.
We then augment $\mathbf{X}$ with an \texttt{[OBJ]} token:
\begin{equation}
  \mathbf{X}' = [\mathbf{X}; x_{\texttt{[OBJ]}}] \in \mathbb{R}^{(N_x + 1) \times d_x}.
\end{equation}
Since the \texttt{[OBJ]} token serves a similar purpose as the \texttt{[CLS]} token, we initialize $x_{\texttt{[OBJ]}} = \mathbf{X}_{N_x}$.
To regulate the interaction between \texttt{[OBJ]} and the other tokens, we construct a mask $m \in \mathbb{R}^{N_x}$, such that:
\begin{equation}
  \label{eq:mask}
  m_i = \begin{cases}
    \mathbbm{1}\{\text{the }i\text{-th patch overlaps with }p\}, & i < N_x \\
    0,                                                           & i = N_x
  \end{cases}
\end{equation}
where $\mathbbm{1}\{\cdot\}$ is the indicator function.
Intuitively, \cref{eq:mask} means that \texttt{[OBJ]} only attends to the patch tokens that are covered by the original proposal $p$.
To maintain the original attentions among $\mathbf{X}$ in $\mathcal{V}$, our attention mask $\mathbf{M}$ is constructed as follows:
\begin{equation}
  \mathbf{M} = \begin{bmatrix}
    \mathbf{1}^{N_x \times N_x} & \mathbf{0}^{N_x} \\
    m                           & 1                \\
  \end{bmatrix} \in \{0, 1\} ^{(N_x + 1) \times (N_x + 1)}.
\end{equation}

Suppose $\mathcal{V}$ has $L$ attention layers (ignoring FFNs), $\mathcal{V}'$ is defined as follows:
\begin{equation}
  \mathbf{X}^l = \begin{cases}
    \sigma \left(\log\mathbf{M} + \mathbf{Q}^l\left(\mathbf{K}^l\right)^\top\right)\mathbf{V}^l + \mathbf{X}^{l-1}, & 0 < l \le L \\
    \mathbf{X}',                                                                                                    & l = 0
  \end{cases}
\end{equation}
where $\mathbf{Q}^l$, $\mathbf{K}^l$, and $\mathbf{V}^l$ are linear transformations of $\mathbf{X}^{l-1}$, $\sigma$ is Softmax function.
Finally, instead of \texttt{[CLS]}, we take the output of \texttt{[OBJ]} as the proposal embedding, \ie, $\tilde e^O_p = \mathbf{X}^L_{N_x + 1}$.
Iterating over the proposals $\mathcal{P}$, we obtain a set of accurate proposal embeddings $\tilde{\mathcal{E}}^O$.

\subsection{Global and Block Distillation}
\label{sec:global_and_block_distillation}

While the object distillation module $\mathcal{M}^O$ aligns the proposal embeddings $\mathcal{E}^O$ with the optimized CLIP embeddings $\tilde{\mathcal{E}}^O$, the detector lacks a comprehensive understanding of the relation between different proposals.
Therefore, we propose a global distillation module $\mathcal{M}^G$, which transfers the knowledge of the entire image from the CLIP visual encoder $\mathcal{V}$ to the detector:
\begin{eqnarray*}
  e^G &=& f(\text{GAP}(\mathbf{F}_6)) \in \mathbb{R}^d,\\
  \tilde e^G &=& \mathcal{V}(\mathbf{I}) \in \mathbb{R}^d,\\
  \mathcal{L}^G &=& \mathcal{L}_1(e^G, \tilde e^G),
\end{eqnarray*}
where $f(\cdot)$ is a linear transformation function, GAP$(\cdot)$ is the global average pooling, and $\mathbf{I}$ is the input image.

Due to the existence of Human Reporting Bias~\cite{human_reporting_bias}, CLIP is prone to ignore non-salient information in an image, \eg, the background or prominent attributes of objects.
Such information may be valuable for dense prediction tasks like detection~\cite{defeat}.
Thus, we propose a block distillation module $\mathcal{M}^B$ to complement the missing knowledge in $\mathcal{M}^G$.
We evenly divide the input image into several blocks $\mathcal{B} \subset \mathbb{R}^4$ via a partition function $g(\cdot)$ and denote the corresponding image regions as $\mathbf{I}^{\mathcal{B}}$.
The size of each block is fixed to $R \times R$, where $R$ denotes the input resolution of $\mathcal{V}$.
In this way, the resize and center crop operations in $\mathcal{V}$ will not take effect, thus avoiding information loss when using $\mathcal{V}$ to encode the block embeddings $\tilde{\mathcal{E}}^B = \{\tilde e^B_b\}_{b \in \mathcal{B}} \subset \mathbb{R}^d$.

On the student side, we apply block pooling and a block head to extract the block embeddings $\mathcal{E}^B = \{e^O_b\}_{b \in \mathcal{B}} \subset \mathbb{R}^d$.
The block pooling is a combination of the block partition function $g(\cdot)$ and RoI Align.
The loss of our proposed block distillation is defined as:
\begin{equation}
  \mathcal{L}^B = \mathcal{L}_1(\mathcal{E}^B, \tilde{\mathcal{E}}^B).
\end{equation}
Compared with $\mathcal{L}^G$, $\mathcal{L}^B$ distills the knowledge of each block with the same weights.
Therefore, the ignored information in $\mathcal{M}^G$ can be compensated by $\mathcal{M}^B$.
Note that neither $\mathcal{M}^G$ nor $\mathcal{M}^B$ is used during inference.

\subsection{Pseudo Label Generation}
\label{sec:pseudo_label_generation}

To investigate the performance of our OADP framework under the G-OVD benchmark (refer to \cref{sec:ovd_benchmarks}), we propose to generate pseudo labels with our modified $\mathcal{V}'$.
Given a proposal $p$, we first extract the proposal embedding $\tilde e^O_p$ as described in \cref{sec:object_distillation}.
Then, the probability of $p$ belonging to category $c \in \mathcal{C}$ is given as:
\begin{eqnarray}
  l^{\text{PL}}(p, c) &=& \frac
  {\tilde e^O_p \cdot t_c}
  {||\tilde e^O_p|| \cdot ||t_c||},\\
  P^{\text{PL}}_{\mathcal{C}}(p, c) &=& \frac
  {\exp\left(l^{\text{PL}}(p, c)\right)}
  {\sum\limits_{c' \in C} \exp\left(l^{\text{PL}}(p, c')\right)}, \label{eq:pl_softmax}
\end{eqnarray}
where $t_c$ is the category embedding of $c$.
Since $P^{\text{PL}}_{\mathcal{C}}(p, c)$ does not reflect the localization quality of $p$~\cite{vild}, we define the confidence score $S_{\mathcal{C}}(p, c)$ as:
\begin{equation}
  S_{\mathcal{C}}(p, c) = P^{\text{PL}}_{\mathcal{C}}(p, c)^\gamma \cdot o_p^{(1 - \gamma)},
\end{equation}
where $\mathcal{C}$ includes both base and novel categories, $o_p \in [0, 1]$ is the objectness score of $p$, and $\gamma$ is a constant balancing factor.
$S_{\mathcal{C}}(p, c)$ reflects the probability that $p$ precisely locates an instance of category $c$.
Finally, we apply class-wise NMS on the novel categories to obtain the pseudo labels.

Note that the Softmax operation in eq.~(\ref{eq:pl_softmax}) is performed over all categories $\mathcal{C}$, even though the pseudo labels do not include instances on base categories.
Such a design effectively suppresses false positives in the pseudo labels.

\begin{table}[t]
  \centering
  \resizebox{\columnwidth}{!}{%
    \begin{tabular}{cc|ccc}
      \toprule
      Benchmark               & Method                        & \mAPN{50}                         & \mAPB{50}                                 & \mAP{50}                                  \\ \midrule
                              & SB~\cite{zsd}                 & 0.3                               & \color{Dim Gray} 29.2                     & \color{Dim Gray} 24.9                     \\
                              & DELO~\cite{delo}              & 3.4                               & \color{Dim Gray} 13.8                     & \color{Dim Gray} 13.0                     \\
      \multirow{-3}{*}{ZSD}   & PL~\cite{pl}                  & 4.1                               & \color{Dim Gray} 35.9                     & \color{Dim Gray} 27.9                     \\ \midrule
                              & ViLD~\cite{vild}              & 27.6                              & \color{Dim Gray} 59.5                     & \color{Dim Gray} 51.3                     \\
                              & RegionCLIP*~\cite{regionclip} & 14.2                              & \color{Dim Gray} 52.8                     & \color{Dim Gray} 42.7                     \\
      \multirow{-3}{*}{V-OVD} & OADP (Ours)                   & \cellcolor{Platinum}\textbf{30.0} & \cellcolor{Platinum}\color{Dim Gray} 53.3 & \cellcolor{Platinum}\color{Dim Gray} 47.2 \\ \midrule
                              & OVR-CNN~\cite{ovr_cnn}        & 22.8                              & \color{Dim Gray} 46.0                     & \color{Dim Gray} 39.9                     \\
                              & HierKD~\cite{hierkd}          & 20.3                              & \color{Dim Gray} 51.3                     & \color{Dim Gray} 43.2                     \\
                              & RegionCLIP~\cite{regionclip}  & 26.8                              & \color{Dim Gray} 54.8                     & \color{Dim Gray} 47.5                     \\
                              & LocOV~\cite{locov}            & 28.6                              & \color{Dim Gray} 51.3                     & \color{Dim Gray} 45.7                     \\
      \multirow{-5}{*}{C-OVD} & PB-OVD~\cite{pb_ovd}          & 29.1                              & \color{Dim Gray} 44.4                     & \color{Dim Gray} 40.4                     \\ \midrule
                              & OV-DETR~\cite{ov_detr}        & 29.4                              & \color{Dim Gray} 61.0                     & \color{Dim Gray} 52.7                     \\
                              & VL-PLM~\cite{vl_plm}          & 32.3                              & \color{Dim Gray} 54.0                     & \color{Dim Gray} 48.3                     \\
      \multirow{-3}{*}{G-OVD} & OADP (Ours)                   & \cellcolor{Platinum}\textbf{35.6} & \cellcolor{Platinum}\color{Dim Gray} 55.8 & \cellcolor{Platinum}\color{Dim Gray} 50.5 \\ \midrule
                              & WSDNN~\cite{wsdnn}            & 19.7                              & \color{Dim Gray} 19.6                     & \color{Dim Gray} 19.6                     \\
      \multirow{-2}{*}{WSD}   & Cap2Det~\cite{cap2det}        & 20.3                              & \color{Dim Gray} 20.1                     & \color{Dim Gray} 20.1                     \\ \midrule
      WS-OVD                  & Detic~\cite{detic}            & 27.8                              & \color{Dim Gray} 47.1                     & \color{Dim Gray} 45.0                     \\ \bottomrule
    \end{tabular}%
  }
  \caption{
    Comparison with other state-of-the-art methods on the OV-COCO dataset.
    Methods are grouped by the benchmark they use.
    ``ZSD'' and ``WSD'' stand for Zero-Shot Detection and Weakly Supervised Detection.
    ``V-OVD'', ``C-OVD'', ``G-OVD'', and ``WS-OVD'' are introduced in \cref{sec:ovd_benchmarks}.
    ``RegionCLIP*'' indicates a model without refinement using COCO-Captions.
  }
  \label{tab:coco_sota}
\end{table}

\begin{table}[t]
  \centering
  \resizebox{\columnwidth}{!}{%
    \begin{tabular}{c|cccc|cccc}
      \toprule
                               & \multicolumn{4}{c|}{Object Detection} & \multicolumn{4}{c}{Instance Segmentation}                                                                                                                                                                                                                                                                 \\
      \multirow{-2}{*}{Method} & \APr                                  & \APc                                      & \APf                                      & \AP                                       & \APr                              & \APc                                      & \APf                                      & \AP                                       \\ \midrule
      ViLD~\cite{vild}         & 16.7                                  & \color{Dim Gray} 26.5                     & \color{Dim Gray} 34.2                     & \color{Dim Gray} 27.8                     & 16.6                              & \color{Dim Gray} 24.6                     & \color{Dim Gray} 30.3                     & \color{Dim Gray} 25.5                     \\
      DetPro~\cite{detpro}     & 20.8                                  & \color{Dim Gray} 27.8                     & \color{Dim Gray} 32.4                     & \color{Dim Gray} 28.4                     & 19.8                              & \color{Dim Gray} 25.6                     & \color{Dim Gray} 28.9                     & \color{Dim Gray} 25.9                     \\
      OV-DETR~\cite{ov_detr}   & -                                     & \color{Dim Gray} -                        & \color{Dim Gray} -                        & \color{Dim Gray} -                        & 17.4                              & \color{Dim Gray} 25.0                     & \color{Dim Gray} 32.5                     & \color{Dim Gray} 26.6                     \\
      OADP (Ours)              & \cellcolor{Platinum}\textbf{21.9}     & \cellcolor{Platinum}\color{Dim Gray} 28.4 & \cellcolor{Platinum}\color{Dim Gray} 32.0 & \cellcolor{Platinum}\color{Dim Gray} 28.7 & \cellcolor{Platinum}\textbf{21.7} & \cellcolor{Platinum}\color{Dim Gray} 26.3 & \cellcolor{Platinum}\color{Dim Gray} 29.0 & \cellcolor{Platinum}\color{Dim Gray} 26.6 \\ \bottomrule
    \end{tabular}%
  }
  \caption{
    Comparison with other state-of-the-art methods on the OV-LVIS dataset.
  }
  \label{tab:lvis_sota}
  \vspace{-6pt}
\end{table}

\begin{table*}[t]
  \centering
  \resizebox{0.9\textwidth}{!}{%
    \begin{tabular}{@{}ccc|ccc|ccc|ccc@{}}
      \toprule
      Global       & Block        & Object       & \multicolumn{3}{c|}{Novel} & \multicolumn{3}{c|}{Base} & \multicolumn{3}{c}{All}                                                                                                                                                                               \\
      Distillation & Distillation & Distillation & \mAP{}                     & \mAP{50}                  & \mAP{75}                & \mAP{}                     & \mAP{50}                   & \mAP{75}                   & \mAP{}                     & \mAP{50}                   & \mAP{75}                   \\ \midrule
                   &              &              & 13.32                      & 24.99                     & 12.35                   & \color{Spanish Gray} 31.87 & \color{Spanish Gray} 50.29 & \color{Spanish Gray} 34.03 & \color{Spanish Gray} 27.02 & \color{Spanish Gray} 43.67 & \color{Spanish Gray} 28.36 \\
      \checkmark   &              &              & 13.51                      & 25.72                     & 12.36                   & \color{Spanish Gray} 32.82 & \color{Spanish Gray} 51.89 & \color{Spanish Gray} 35.31 & \color{Spanish Gray} 27.77 & \color{Spanish Gray} 45.04 & \color{Spanish Gray} 29.31 \\
                   & \checkmark   &              & 14.57                      & 27.25                     & 13.17                   & \color{Spanish Gray} 34.45 & \color{Spanish Gray} 53.60 & \color{Spanish Gray} 37.20 & \color{Spanish Gray} 29.25 & \color{Spanish Gray} 46.71 & \color{Spanish Gray} 31.06 \\
                   &              & \checkmark   & 15.49                      & 27.23                     & 15.25                   & \color{Spanish Gray} 35.99 & \color{Spanish Gray} 55.96 & \color{Spanish Gray} 38.57 & \color{Spanish Gray} 30.63 & \color{Spanish Gray} 48.45 & \color{Spanish Gray} 32.47 \\ \midrule
      \checkmark   & \checkmark   &              & 13.50                      & 26.49                     & 12.50                   & \color{Spanish Gray} 32.19 & \color{Spanish Gray} 51.25 & \color{Spanish Gray} 33.94 & \color{Spanish Gray} 27.30 & \color{Spanish Gray} 44.78 & \color{Spanish Gray} 28.33 \\
      \checkmark   &              & \checkmark   & 15.47                      & 28.80                     & 14.62                   & \color{Spanish Gray} 34.08 & \color{Spanish Gray} 54.29 & \color{Spanish Gray} 36.28 & \color{Spanish Gray} 29.21 & \color{Spanish Gray} 47.62 & \color{Spanish Gray} 30.61 \\
                   & \checkmark   & \checkmark   & 15.92                      & 29.01                     & \textbf{15.64}          & \color{Spanish Gray} 35.30 & \color{Spanish Gray} 55.45 & \color{Spanish Gray} 37.88 & \color{Spanish Gray} 30.23 & \color{Spanish Gray} 48.53 & \color{Spanish Gray} 32.06 \\
      \checkmark   & \checkmark   & \checkmark   & \textbf{16.21}             & \textbf{29.95}            & 15.47                   & \color{Spanish Gray} 33.33 & \color{Spanish Gray} 53.26 & \color{Spanish Gray} 35.47 & \color{Spanish Gray} 28.85 & \color{Spanish Gray} 47.17 & \color{Spanish Gray} 30.24 \\ \bottomrule
    \end{tabular}%
  }
  \caption{
    Ablation study of the Global, Block, and Object Distillation modules in the OADP framework.
    The baseline is our re-implemented ViLD-ensemble model.
  }
  \label{tab:oadp}
  \unless\ifarxiv
    \vspace{-12pt}
  \fi
\end{table*}

The proposals to be labeled are extracted via an RPN model pretrained on $\mathcal{D}^T$, which only contains annotations of the base categories.
ViLD~\cite{vild} demonstrates that the generalization ability of $\mathcal{P}$ is strong enough to recall most objects of the novel categories.

\section{Experiments}

In this section, we first introduce the detailed experiment setup, including the datasets, evaluation metrics, and implementation details.
We then evaluate the performance of our proposed OADP framework and analyze the results compared to the state-of-the-art approaches.

\subsection{Datasets}

Experiments are mainly conducted under the open-vocabulary COCO (OV-COCO) setting~\cite{ovr_cnn}, where the MS-COCO 2017 dataset~\cite{coco} is manually divided into $48$ base categories and $17$ novel categories.
The training dataset contains $107,761$ images, and the validation dataset contains $4,836$ images.
We report the \mAPN{50}, \mAPB{50}, and \mAP{50} metrics, \ie, the mAP at IoU threshold $0.5$ for novel, base, and all categories.
\mAPN{} is the main metric.
Some experiments are conducted under the open-vocabulary LVIS (OV-LVIS)~\cite{vild} setting, where the $337$ rare categories in LVIS~\cite{lvis} are treated as novel categories, and the other $866$ are base categories.
Metrics for the OV-LVIS setting are \APr, \APc, \APf, and \AP, \ie, the mAP for rare (novel), common, frequent, and all categories.
Both object detection and instance segmentation metrics are reported.

\subsection{Implementation Details}

Training is conducted on $8$ V-100 GPUs with batch size $16$ in total.
We use stochastic gradient descent (SGD) optimizer with $0.02$ initial learning rate, $0.9$ momentum, and $2.5 \times 10^{-5}$ weight decay.
The student backbone is ResNet-$50$~\cite{resnet}.
Following DetPro~\cite{detpro}, we adopt the ViT-B/32 CLIP~\cite{clip} as the teacher and initialize the student backbone using SoCo~\cite{soco}.
The loss weights $w^O$, $w^B$, and $w^G$ are set to $0.5$, $0.25$, and $0.25$ respectively.
Under the OV-COCO setting, we train the detector for $40,000$ iterations.
At the $32,000$\textsuperscript{th} iteration, the learning rate is divided by 10.
For OV-LVIS, we use $2$x ($24$ epochs) training schedule, where the learning rate is divided by $10$ at the $16$\textsuperscript{th} and $22$\textsuperscript{th} epochs.

\subsection{Main Results}

We compare our OADP framework with the other state-of-the-art OVD methods.
As described in \cref{sec:ovd_benchmarks}, we categorize existing OVD methods by the benchmark they belong to.
For completeness, we include two related benchmarks: Zero-Shot Detection (ZSD)~\cite{zsd, Rahman2020, gtnet, Huang2021a, Li2019c} and Weakly Supervised Detection (WSD)~\cite{wsdnn, cap2det}.

Our method mainly focuses on the V-OVD benchmark and the G-OVD benchmark.
As shown in \cref{tab:coco_sota}, our OADP framework achieves $30.0$ \mAPN{50} on the V-OVD benchmark.
RegionCLIP*~\cite{regionclip} uses CLIP~\cite{clip} as the pretrained weight, thus adhering to the V-OVD benchmark.
Some V-OVD methods~\cite{glip, mdetr, x_detr, owl_vit, detclip} are not included because they rely on large-scale detection and image-text datasets and cannot be compared fairly.
While C-OVD is not our primary concern, we include the corresponding methods for reference and report their performance when only COCO Captions~\cite{coco_captions} is available.
Under such constraint, the performance of the caption-based methods is relatively lower than the V-OVD methods, even if additional caption data is used.
Moreover, our OADP framework is perpendicular to the caption-based methods and has the potential to achieve higher performance using captions.

For the G-OVD benchmark, we generate pseudo labels for novel categories as described in \cref{sec:pseudo_label_generation}.
The pseudo labels are then merged with the instance-level annotations for base categories.
Training our OADP framework on the mixed dataset yields $35.6$ \mAPN{50}, surpassing the previous SOTA method VL-PLM~\cite{vl_plm} by $3.3$ \mAPN{50}.
Since PromptDet~\cite{promptdet} relies on an external dataset (LAION-400M~\cite{laion_400m}) and uses smaller image size ($640 \times 640$), the result $26.6$ \mAPN{50} is not listed in \cref{tab:coco_sota} for fairness.

\Cref{tab:lvis_sota} shows the comparison between our method and the other state-of-the-art methods on the OV-LVIS dataset.
Most of the methods under the OV-LVIS setting adhere to the V-OVD benchmark, so we conduct experiments on the V-OVD benchmark only.
For the object detection task, our OADP framework achieves $21.9$ \APr, surpassing DetPro~\cite{detpro} by $1.1$ \APr.
We also report the performance of the instance segmentation task, which achieves $21.7$ \APr{} and is $1.9$ \APr{} higher than the previous SOTA method.

\subsection{Ablation Study}

We conduct ablation studies on the OV-COCO dataset to evaluate the effectiveness of each component in our proposed OADP framework.

\paragraph{OADP.} \Cref{tab:oadp} shows the effectiveness of each distillation module in our OADP framework.
The first row is our re-implemented ViLD-ensemble~\cite{vild}.
Due to the expensive training cost of ViLD, the performance $24.99$ \mAPN{50} is far below the official $27.60$ \mAPN{50}.
Nevertheless, with our proposed distillation pyramid, we are able to surpass ViLD eventually.
The $2$\textsuperscript{nd} to $4$\textsuperscript{th} row in \cref{tab:oadp} adds $\mathcal{M}^G$, $\mathcal{M}^B$, and $\mathcal{M}^O$ to the baseline respectively.
The global distillation module brings a $0.73$ \mAPN{50} gain, while the other two bring $2.26$ \mAPN{50} and $2.24$ \mAPN{50} gain.
Note that by adding $\mathcal{M}^O$ to the baseline, we remove the original image head in ViLD-ensemble.
Therefore, the $2.24$ \mAPN{50} gain is a result of the OAKE module instead of the distillation operation.
The $5$\textsuperscript{th} row in \cref{tab:oadp} adds $\mathcal{M}^G$ and $\mathcal{M}^B$ together.
While the $26.49$ performance is higher than sole $\mathcal{M}^G$, it is slightly lower than the $27.25$ \mAPN{50} of $\mathcal{M}^B$.
However, along with the object distillation module $\mathcal{M}^O$, $\mathcal{M}^G$ and $\mathcal{M}^B$ achieves $28.80$ \mAPN{50} and $29.01$ \mAPN{50}, suggesting that $\mathcal{M}^G$ and $\mathcal{M}^B$ have a similar function in transferring the global scene knowledge from CLIP to the detector.
Finally, using all three modules together, we achieve $29.95$ \mAPN{50}.

\begin{table}[t]
  \centering
  \resizebox{\columnwidth}{!}{%
    \begin{tabular}{@{}c|cc|cc@{}}
      \toprule
      \multirow{2}{*}{Method} & \multicolumn{2}{c|}{Macro Precision} & \multicolumn{2}{c}{Weighted Precision}                             \\
                              & w/o OAKE                             & w/ OAKE                                & w/o OAKE & w/ OAKE        \\ \midrule
      Baseline                & 58.08                                & -                                      & 62.04    & -              \\
      ViLD*                   & 63.36                                & -                                      & 65.91    & -              \\
      MBS                     & 61.70                                & 63.83                                  & 64.81    & 65.82          \\
      Fixed                   & 49.07                                & 64.53                                  & 51.49    & \textbf{69.75} \\
      Adaptive                & 51.64                                & \textbf{66.09}                         & 55.85    & 68.68          \\ \bottomrule
    \end{tabular}
  }
  \caption{
    Ablation study of OAKE module.
    ``ViLD*'' indicates our re-implementation of multi-scale region embedding.
    ``MBS'', ``Fixed'', and ``Adaptive'' are three transforming strategies.
  }
  \label{tab:token}
\end{table}

\paragraph{OAKE.} We demonstrate the effectiveness of our OAKE module in \cref{tab:token}.
Given the ground truth bounding boxes, we use different strategies to crop their image regions.
(1) Baseline: $1\times$ crop;
(2) ViLD*: $1\times$ and $1.5\times$ crop;
(3) MBS: the minimum bounding square of the original bounding box;
(4) Fixed: $224\times 224$ bounding square;
(5) Adaptive: adaptively enlarge the bounding square.
For the above strategies, we use CLIP to directly extract embeddings for their image regions (``w/o mask'').
Alternatively, we can utilize the modified CLIP visual encoder mentioned in~\cref{sec:object_distillation} (``w/ mask'').
Finally, we classify these embeddings by calculating their similarity with category embeddings.
To evaluate the performance, we compute ``Macro Precision'' (precision for each category independently with equal weights) and ``Weighted Precision'' (weights depending on the number of bounding boxes in each class).

\begin{figure*}[t]
  \centering
  \includegraphics[width=\linewidth]{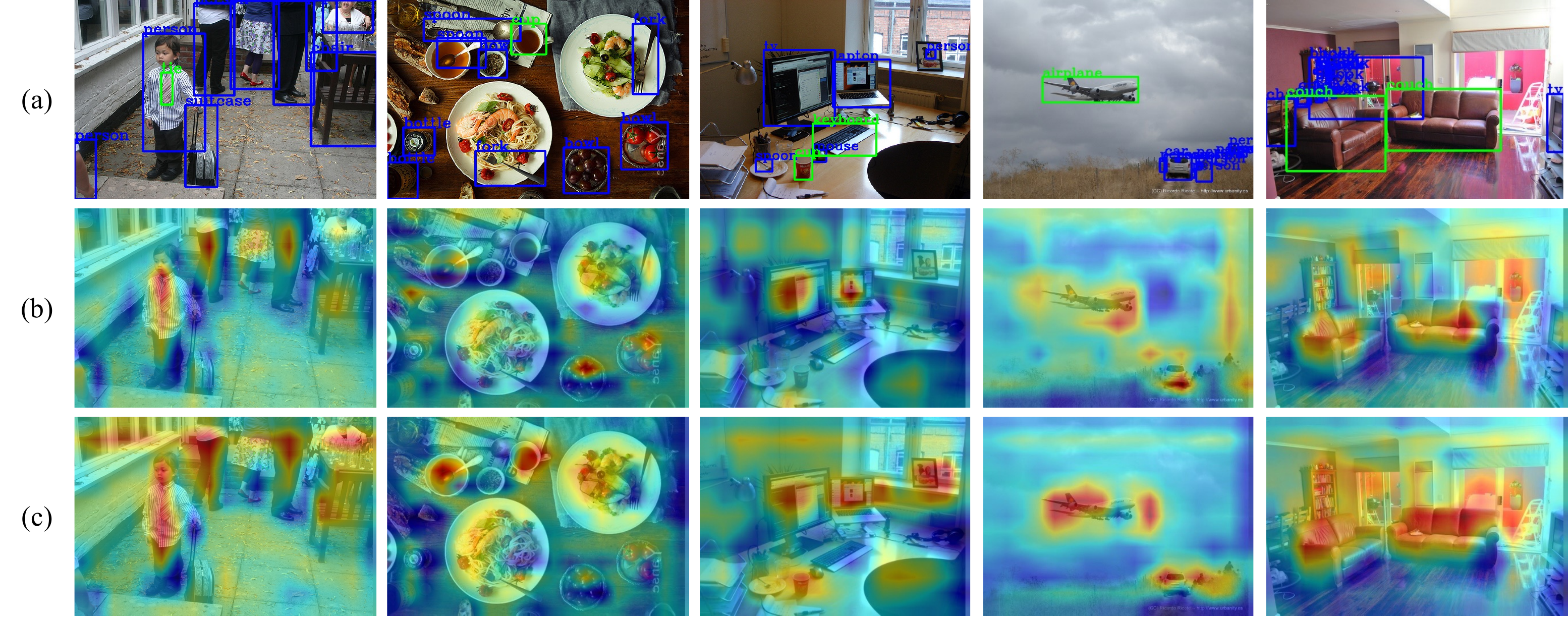}
  \caption{
    Visualization of activation patterns from different detectors.
    (a) Pseudo labels (green) and ground truth annotations (blue) for each image.
    (b) Baseline detector.
    (c) OADP detector.
    The intensity of the feature response increases from blue to red.
  }
  \label{fig:vis}
\end{figure*}

As shown in the $1$\textsuperscript{st} and $3$\textsuperscript{rd} columns, ``Fixed'' and ``Adaptive'' strategies bring performance drops as they crop a larger bounding square compared to other strategies (\eg, ``MBS'') which may introduce additional surrounding distractors that confuse the CLIP visual encoder.
However, with our object-aware CLIP visual encoder, the performances of the two strategies boost significantly and suppress others.
It validates that the CLIP visual encoder can focus on the proposal object with our mask attention mechanism to extract accurate knowledge.

\begin{table}[t]
  \centering
  \begin{tabular}{c|cccc}
    \toprule
    Method               & \mAPPL{}        & \mAPPL{50}      & \mAPPL{75}      & \#PL \\ \midrule
    Baseline             & \hphantom{0}5.0 & 12.1            & \hphantom{0}3.1 & 100  \\
    ViLD*                & 17.8            & 28.1            & 18.6            & 100  \\
    Ours                 & \textbf{19.0}   & \textbf{29.9}   & \textbf{19.9}   & 100  \\ \midrule
    Baseline             & \hphantom{0}3.9 & \hphantom{0}9.3 & \hphantom{0}2.5 & 6.53 \\
    VL-PLM~\cite{vl_plm} & -               & 25.3            & -               & 4.26 \\
    Ours                 & \textbf{17.4}   & \textbf{26.5}   & \textbf{18.6}   & 4.14 \\ \bottomrule
  \end{tabular}
  \caption{
    Ablation study of pseudo labels.
    ``\#PL'' is the number of pseudo labels per image.
  }
  \label{tab:pseudo_label}
  \vspace{-12pt}
\end{table}

\paragraph{Pseudo Label.} We follow VL-PLM\cite{vl_plm} to adopt the COCO-ZS setting for our ablation studies of the pseudo label.
Both mAP and the average per-image number of PLs (\#PL) on novel categories are used as metrics to evaluate the quality of the pseudo label.
The baseline method directly uses \texttt{[CLS]} token of CLIP visual encoder to extract proposal embeddings from the original proposal, and it relies merely on the classification score to sort proposals.
The poor detection accuracy in $1^{\text{st}}$ and $4^{\text{th}}$ rows of \cref{tab:pseudo_label} show that without the objectness score, the baseline method can not accurately localize objects.
We re-implement multi-scale region embedding of ViLD\cite{vild} with a geometric mean of CLIP classification score and objectiveness score, \ie, ``ViLD*''.
We adopt an adaptive transform strategy for proposals and regard the output of \texttt{[OBJ]} as proposal embedding.
The score fusion strategy is described in \cref{sec:pseudo_label_generation}, where $\gamma$ is 0.3.
Our method achieves the highest mAP when the number of PLs is sufficient.
VL-PLM\cite{vl_plm} adopts a multi-scale region embedding method similar to ViLD\cite{vild} except for an arithmetic mean of classification score and objections score.
When the pseudo labels are filtered with a higher confidence threshold, our method still has a significant advantage compared to VL-PLM\cite{vl_plm} ($26.50$ \mAPPL{50} compared to $25.30$ \mAPPL{50}).

\subsection{Visualization}

We visualize our generated PLs on novel categories in green with ground truth boxes of base categories in blue (\cref{fig:vis} (a)).
We try our best to ensure the accuracy of PLs as much as possible to be fewer and more precise.
These green PLs demonstrate that our proposal embeddings can clearly distinguish novel objects from base ones.
Correspondingly, we also show activation maps from baseline (b) and our detector (c) in \cref{fig:vis}.
Taking the $2$\textsuperscript{nd} column as an example, the activation map of our detector accurately highlights more area of novel objects, \ie ``cup'', with our distillation pyramid mechanism.
Therefore, the backbone of OADP generates
more informative feature maps, which further help detect novel objects.

\section{Conclusion}

In this paper, we reconsider the way of knowledge extraction and knowledge transfer in existing KD-based OVD methods and propose an \OADP{} (OADP) framework.
To preserve complete and purified object representation in proposals during knowledge extraction, we propose an \OAKE{} (OAKE) module to adaptively transform proposals and extract precise object knowledge.
A \DP (DP) mechanism is proposed to transfer contextual knowledge about the relation of different objects for better scene understanding.
Experiments show that our OADP outperforms previous methods on two popular OVD benchmarks.

{
    \small
    \bibliographystyle{ieee_fullname}
    \bibliography{library}
}

\end{document}